# Data Collection and Analysis of French Dialects

**Omar Shaur Choudhry, Paul Omara Odida, Joshua Reiner,**
**Keiron Appleyard, Danielle Kushnir** and **William Toon**
University of Leeds, LS2 9JT, United Kingdom
{sc20osc,sc20poo,sc20jdr,ed19k3a,sc20dk,sc20wt}@leeds.ac.uk

## Abstract

*This paper discusses creating and analysing a new dataset for data mining and text analytics research, contributing to a joint Leeds University research project for the Corpus of National Dialects. This report investigates machine learning classifiers to classify samples of French dialect text across various French-speaking countries. Following the steps of the CRISP-DM methodology, this report explores the data collection process, data quality issues and data conversion for text analysis. Finally, after applying suitable data mining techniques, the evaluation methods, best overall features and classifiers and conclusions are discussed.*

## 1 Introduction

English is an international language used across multiple countries all over the world. The International Corpus of English includes samples of different dialects of English from many different regions (Atwell, 2021). Whilst there are many other modern languages used both natively and as a second language, there are no equivalent International Corpora for non-English languages (Atwell, 2021). In this report, we investigate samples of national dialects of French across Algeria, the Democratic Republic of Congo, France, Ivory Coast, Morocco and Senegal.

This study implements the Cross-Industry Standard Process for Data Mining (CRISP-DM) methodology. Software used in this project includes proprietary tools such as SketchEngine and WebBootCat and an open-source tool WEKA (Waikato Environment for Knowledge Analysis).

### 1.1 Requirements

The main requirement of this project is to collate subcorpora each between 50,000 and 70,000 words, containing samples of national dialects of a non-English language across different countries, and apply data mining and text analytics tools to identify distinguishable features of each national dialect, experimenting with machine learning classifiers to classify samples of dialect text. Further requirements include investigating theory, methods and terminology used in data mining and text analytics; experiencing how to apply algorithms, resources and techniques for implementing and evaluating data mining and text analytics in a practical research exercise; and summarising and presenting achievements to a peer audience in a research conference paper.

### 1.2 Business Objectives

This report has chosen to investigate French dialects across various French-speaking countries. Therefore, the primary business goal of this project is to create and analyse a new data set, consisting of samples of national dialects of French across various French-speaking countries.

### 1.3 Data Mining Problem Definition

CRISP-DM is the methodology applied to achieve this report's business objectives. The main data mining problem definition is classification. Firstly, data sets for each national sub-corpora are created. Secondly, appropriate data mining tools are used to classify samples of these data sets by their respective national dialect. Thirdly, we investigate potential distinguishable features.

## 2 Data Understanding

### 2.1 Data Collection Process

To create each national subcorpus, we used the WebBootCat feature in SketchEngine to retrieve texts from the web with the specific top-level domain (TLD) of the respective country using 10 randomly chosen seed search terms from the Leeds Internet Corpus Website (Sharoff, 2006).

### 2.2 Data Quality Issues

There were various issues regarding data quality.

In the Algerian subcorpus, artefacts of Arabic text remained. We kept these to preserve the integrity of what the language looks like, even if it contains some multi-word expressions (MWEs) of Arabic French.

Another issue regarding SketchEngine data collection is that WebBootCat can cut out portions of text, mistaking it for boilerplate (Al-Sulaiti et al., 2016). Accessing certain web pages used to generate the corpus, not all text was collected.

We also found that using 5 seed terms was far too small to generate sufficiently sized corpora. Using 9 terms led to some corpora being too small whilst using 10 terms led to some corpora being too large, thus we had to manually remove documents to maintain the subcorpus word limit.

Some documents were far too large, thus we set a 3000-word limit for a single document to remove any large documents which might have otherwise created significant bias.

Initially, we used different seed terms, however this caused problems. The seed terms were too specific, relating to politics and COVID-19, leading to overly specific corpora. Furthermore, specific genres of websites contain similar language (Asheghi, Markert and Sharoff, 2014), so selecting seed terms around a certain genre would search for specific genres of websites, which could affect accuracy when modelling. Moreover, certain genres may not be as prevalent across different TLDs, which could skew the language used and should be avoided.

## 3 Data Preparation

We can use the explorer feature in WEKA to train machine learning classifiers to classify samples of dialect text.

### 3.1 Dataset

WEKA requires the input file to be in Attribute-Relation File Format (ARFF) – an ASCII text file that describes a list of instances sharing a set of attributes (Paynter, 2008).

We set the relation to "french" and set an attribute "text" - a string containing the sample of text, with each instance labelled with its national TLD as its "class" used as a class nominal label.

To investigate our dataset, we created two separate ARFF files; one used entire documents that were collected using SketchEngine as instances and the other used tokenization in pre-processing to create individual text samples.

### 3.2 Data Cleaning and Pre-Processing

We parsed and concatenated the individual subcorpus using custom Python scripts with the header and format required for the ARFF files.

In the first file, we formatted each instance in the ARFF format ["text", TLD] where "text" is a string containing the instance (removing the actual document tags) and TLD is the relative top-level domain of the sub-corpus country, respectively as the class label for each instance in the relative subcorpus.

To generate the second file, we again removed any document, "<p>" and "</p>"(since there were some issues where these tags were missing in some documents, they were removed entirely from all instances) tags generated from the initial corpus creation; and similarly, formatted each instance in the ARFF format.

For tokenization, we used the Tokenizer package in the Natural Language Toolkit, an open-source Python library for Natural Language Processing, to divide the sample text using the French Punkt sentence tokenizer. This divided paragraphs of text into a list of sentences using an unsupervised algorithm to build a model for abbreviation words, collocations and words that start sentences (NLTK Project, 2022).

## 4 Modelling

As previously stated, the data mining objective of this project was not to build a highly accurate model, but rather to apply various data mining and text analytics techniques to investigate data mining theory, methods and terminology. Therefore, it was decided to perform classification across various features and classifiers. All processing was executed in WEKA.

### 4.1 Features

Initially, we tested the entire corpus text as features, without any further filtering, with the document ARFF file slightly outperforming the tokenized ARFF file. In this case, many classifiers were incompatible. Thus, we applied the **String ToWordVector** (STWV) filter (Trigg et al., 2008), by extracting corpus text words as features.

We further filtered the data by using the **Ranker** search method within the **Information Gain Attribute Evaluator** through the **Attribute Selection** filter to reduce the dataset to improve classifier performance by measuring information gained concerning the class, by ranking attributes by their evaluations. We tested different threshold values by which attributes can be discarded. Now, features which have little information gain can be removed. Some of the features with the highest information values include "*Sénégal*", "*Côte*", "*Casablanca*" and "*Kinshasa*" which are all geographical locations. It is likely these features that can produce accurate models but will reflect on the country of origin, not the actual dialects and may even perhaps be identified as distinguishable features by certain classifiers.

We also investigated filtering out commonly used stop words, which could be wrongly identified as distinguishable features, dominating an analysis but not offering much insight into the classification of the dialect text (Savoy, n.d.).

### 4.2 Classification

Testing multiple classifiers across a range of features can be used to prevent inconsistencies and can help handle edge cases (Di Bari, Sharoff and Thomas, 2014). In total, we tested 5 classifiers: **Multinomial Naive Bayes** – determines the conditional probability of events, **Logistic** – a multinomial logistic regression model with a ridge estimator, **SMO** – WEKA's implementation of the Support Vector Machines classifier (Alshutayri et al. 2016), **Bagging** – creates and combines results from separate samples and **J48** – build decision trees by splitting features which contain the most information.

### 4.3 Results

To test the features and classifiers we must first describe the test methods we can use for testing models; training set – using the same data used to train the classifier as test data; cross-validation – a systematic way of improving upon repeated holdout which improves by reducing the variance of the estimate; and percentage split – using a certain percentage of the ARFF file as training data and the rest as test data. Table 1 reports the average value of correctly identified instances by measuring the performance of classifiers testing all features across all test methods listed.

| Classifier | Accuracy |
|---|---|
| NaiveBayesMultinomial | 66.30 |
| Logistic | 64.64 |
| SMO | 73.11 |
| Bagging | 79.83 |
| J48 | **81.33** |

Table 1: Average Accuracy of Classifiers (%)

Table 2 reports the average value of correctly identified instances by measuring the performance of features of all classifiers across all test methods. As the Ranker threshold was increased, the accuracy was reduced, as there were fewer features the classifier used to generate the model.

| Feature | Accuracy |
|---|---|
| StringToWordVector | 72.23 |
| StopWordsHandler | 73.49 |
| AttributeSelection (0 Threshold) | **84.17** |
| AttributeSelection (0.05 Threshold) | 76.27 |
| AttributeSelection (0.1 Threshold) | 80.67 |
| AttributeSelection (0 Threshold) + StopWordsHandler | 79.27 |

Table 2: Average Accuracy of Filters (%)

In testing the 2 ARFF files, we were not able to generate very accurate results when using the tokenized dataset. The tokenized data set contained too many samples, thus creating models that would either overfit or not classify text as accurately as the non-tokenized data set. Furthermore, many tests took a very long time to run. Since samples were smaller, the classification models may not be as accurate either, thus we did perform many tests with these.

### 4.4 Other Methods

Through the corpora comparison feature in SketchEngine, we investigated the similarity of the corpora against each other and with 2 large reference corpora, frTenTen12 and frTenTen17, as demonstrated in Figure 1.

We investigated the most common MWEs of each subcorpus using SketchEngine. We found that "*de la*" was the most common MWE for every corpus. The most common MWEs were combinations of stop words. Since many of these are also the same across each subcorpus, the classification may not be greatly affected when not filtering out stop words. Noticeable MWEs include names of people, such as "*Félix Tshisekedi*" for the Democratic Republic of Congo, and "*Barthélemy Dias*" and "*Ousmane Sonko*" for Senegal. These aren't features which are specific to the French dialect, but rather MWEs specific to the country.

|  | (frTenTen12) | (frTenTen17) | FR-CD | FR-CI | FR-DZ | FR-FR | FR-MA | FR-SN |
|---|---|---|---|---|---|---|---|---|
| (frTenTen12) | 1.00 | 1.47 | 2.86 | 2.57 | 2.60 | 2.37 | 2.12 | 2.53 |
| (frTenTen17) | 1.47 | 1.00 | 2.79 | 2.56 | 2.64 | 2.43 | 2.17 | 2.55 |
| FR-CD | 2.86 | 2.79 | 1.00 | 3.37 | 3.47 | 3.43 | 3.06 | 3.26 |
| FR-CI | 2.57 | 2.56 | 3.37 | 1.00 | 3.05 | 3.12 | 2.83 | 3.03 |
| FR-DZ | 2.60 | 2.64 | 3.47 | 3.05 | 1.00 | 3.07 | 2.72 | 3.13 |
| FR-FR | 2.37 | 2.43 | 3.43 | 3.12 | 3.07 | 1.00 | 2.67 | 3.09 |
| FR-MA | 2.12 | 2.17 | 3.06 | 2.83 | 2.72 | 2.67 | 1.00 | 2.70 |
| FR-SN | 2.53 | 2.55 | 3.26 | 3.03 | 3.13 | 3.09 | 2.70 | 1.00 |

Figure 1: Corpus comparison table

# 5 Evaluation

## 5.1 Evaluation Methods

A problem with using the test set we created is that when using AttributeSelection as a feature, as some words are missing from the model, it does not correctly identify instances as much as with other methods. A problem with using a percentage split is that sometimes the test set split is not representative of the training set split. Cross-validation can generate an overall better model, but it can take much longer than other methods.

Some classifiers used stop words within the analysis, such as the decision tree in J48, when filtering stop words, it produced less accurate results. Nevertheless, when combined with AttributeSelection, J48 outperforms many other algorithms as it can produce more accurate decision trees.

To evaluate features and classifiers, we took the best performing features on average based on a variety of tests through all classifiers. To look for distinguishable features, we looked at the decision trees produced at J48 and the word probabilities per corpus by Naïve Bayes. Even though the average accuracy for NaiveBayesMultinomial is low, it was the only classifier that worked somewhat efficiently for the tokenized data set and thus skewed some of the results.

## 5.2 Best Features and Classifiers

The document ARFF file produced more accurate models than the tokenized ARFF file, thus using larger documents as features was more effective. Based on the results from Table 2, we can assume that the best features to investigate are those once the dataset is filtered through STWV and Attribute Selection with a 0 threshold Ranker. Table 3 reports the average value of correctly identifies instances of the non-tokenized dataset by measuring the performance of each classification model using this filter combination across the specific methods, evaluating the training set (column 2), 10-fold cross-validation (column 3) and percentage split 60% train, 40% test (column 4), a similar method as seen in Alshutayri et al. (2016).

| Classifier | TS | CV | PS |
|---|---|---|---|
| NaiveBayesMultinomial | 84.23 | **77.43** | **78.95** |
| Logistic | **99.2** | 64.83 | 63.82 |
| SMO | 97.38 | 72.97 | 65.79 |
| Bagging | 81.10 | 64.04 | 65.79 |
| J48 | 87.4**0** | 70.87 | 72.37 |

Table 3: Average Accuracy of Classifiers using STWV + Attribute Selection (0) (%)

Based on results from Table 1, we can see, as expected the performance from J48. Even though the Logistic classifier has an accuracy of 99.2% on the training set, however, this is most likely due to overfitting, as it performed much worse in cross-validation and percentage split. We can conclude that from Table 3, the classifiers may be overfitting the training data.

## 5.4 Distinguishable Features

Furthermore, based on the J48 decision tree, we can conclude certain features that can distinguish it from other variants. The tree contained 83 nodes, over half of which were names of cities and countries. As mentioned in subsections 4.1 and 4.4, we found that all algorithms used MWEs of names of people and geographical locations, which is a distinguishable feature, but of the country of origin, not the dialect.

NaiveBayesMultinomial calculated that the feature with the highest probability to appear in any specific corpus text was "*pour"* which is a French stop word. When filtering stop words, classifiers performed better than with just STWV, but not as good as with AttributeSelection (as there are many attributes with negative information gain). The next top 3 probabilities used the "<p>" tags, which again demonstrates the absence of clearly distinguishable features.

## 5.5 Business Issues

Due to the limitations of this report, we were not able to go into great detail in analysing the dataset as much as we would have liked. We could have experimented further with StringToWordVector, testing stemming, word replacement, non-alphabetic character removal, part-of-speech tagging and named entity recognition. Furthermore, applying some more text and data mining methods, such as topic modelling, collocation analysis, and considering document term frequencies would have been interesting, including applying different forms of tokenization than the single one we applied using NLTK such as character n-gram tokenizers.

## 5.6 Meeting Business Objectives

Data mining and text analytics techniques and CRISP-DM methodology have shown we can create and analyse a data set for investigating dialects and variants of a non-English language.

Considering the findings of this project and acknowledging the specifics of this domain, it can be concluded that we have successfully managed to formulate and discuss the business and data mining objectives including the data mining problem definition, creating, understanding and preparing an appropriate dataset suitable for analysis, including pre-processing, to run data mining and text analytics tools to investigate the best features and classifiers, reporting on the evaluation methods.

Furthermore, we have demonstrated how to apply algorithms, resources and techniques for implementing and evaluating data mining and text analytics in a practical research exercise through investigating data mining theory, terminology and methods.

To conclude, we hope that this report helps demonstrate the creation of a data set, consisting of samples of national dialects of French across various French-speaking countries, for data mining and text analytics research and its analysis through different methods, whilst adhering to the CRISP-DM methodology, for the Corpus of National Dialects.